\documentclass[runningheads]{template/gcpr2026_author_kit/llncs}

\RequirePackage{silence}
\usepackage{wrapfig}
\usepackage{graphicx}
\usepackage{comment}
\usepackage{amsmath,amssymb}
\usepackage{color}
\usepackage{url}
\usepackage{hyperref}

\newif\ifreview
\reviewfalse

\ifreview
  \usepackage{lineno}

  \linenumbers
\fi

\input{bib/bib_short.def}

\begin{document}

\def\SubNumber{105}
\def\GCPRTrack{Extended Abstract}

\title{From Laboratory to Road: Evaluating Wearable Gaze Accuracy for Driving}

\ifreview
\else
  \titlerunning{From Laboratory to Road}
  \author{William Engel\inst{1}\orcidID{0009-0008-5184-5624} \and
          Fabian Flohr\inst{1}\orcidID{0000-0002-1499-3790}}
  \authorrunning{W. Engel and F. Flohr}
  \institute{Intelligent Vehicles Lab, Hochschule München, Germany\\
  \email{wengel@hm.edu, fabian.flohr@hm.edu}}
\fi

\maketitle

\begin{abstract}

Bird’s-eye-view (BEV) representations have become a widely used interface between perception and planning in autonomous driving, but they encode what is in a scene, not what is behaviorally relevant to a human driver. Gaze offers a compelling behavioral signal for this gap---yet wearable eye trackers are routinely deployed as if their spatial output were ground truth, despite known sensitivity to head motion, illumination, and calibration drift. We present, to our knowledge, the first unified framework for quantifying wearable gaze accuracy under real driving conditions. Our on-road study contains 41 validated scenes in which one driver fixated a vehicle's license plate. Gaze error is measured as the angular difference between the plate center and the gaze direction estimated by the glasses. Separate indoor studies with the same driver and device systematically analyze how distance, illumination, head motion, target motion, and gaze eccentricity affect both systematic bias and gaze precision. The mean on-road error was 4.58\textdegree. Applying an offset estimated from the indoor recordings reduced it to 1.10\textdegree\ and improved all 41 scenes. Because this offset varied between sessions, reliable BEV supervision may require online recalibration and condition-dependent estimates of gaze uncertainty.

\end{abstract}

\section{Introduction}
\label{sec:introduction}

Autonomous-driving systems increasingly use bird's-eye-view (BEV) representations to combine multi-sensor information in a common spatial frame for perception, prediction, and planning \cite{li2022bevformer,liu2023bevfusion,jiang2023vad}. Although BEV encodes objects, trajectories, and map elements, it does not indicate which scene elements are behaviorally relevant to a human driver.

Driver gaze offers a compact signal of visual selection and has been modeled in datasets such as DR(eye)VE, BDD-A, DADA-2000, and Look Both Ways \cite{palazzi2019dreyeve,xia2018bdda,fang2019dada2000,kasahara2022lookBothWays}. Prior work has used gaze for driving prediction, imitation learning, and object-relevance estimation \cite{liu2019gazeModel,abdelkarim2021ggnet,biswas2024gazeSupervision}. However, wearable gaze cannot be treated as exact spatial supervision without quantifying its accuracy and precision during driving. Calibration drift, head motion, illumination, and pupil dynamics introduce condition-dependent errors \cite{niehorster2026dataQuality,tonsen2020pupilInvisible,hooge2023robust,salari2026pupilSize}; after projection into BEV, even small angular errors may cause large spatial displacements or incorrect object assignments. As gaze is a behavioral sample rather than an exhaustive label, its uncertainty must be represented explicitly.

This pilot study measures wearable gaze accuracy during real driving and asks whether the resulting errors can be explained by matched indoor experiments. On-road targets are declared aloud by the driver and geometrically localized as reference targets. Our framework combines these on-road measurements with five laboratory protocols, each isolating one factor: illumination, head motion, pursuit, eccentricity, and depth. It measures angular accuracy, signed bias, and fixation-level precision, forming a basis for uncertainty-aware gaze supervision in BEV-based driving models.

\section{Related Work}
\label{sec:related-work}

Eye tracking is widely used to study driver attention, situational awareness, and interaction with advanced driver-assistance systems \cite{cvahteOjstersek2019eyeTracking,khan2019gaze,nagy2024takeover}. However, simulator studies provide controlled baselines without reproducing all physical disturbances of real-world driving \cite{calvi2023comparing}.

Validation studies under dynamic conditions identify several relevant error sources. Physical motion can substantially degrade wearable gaze accuracy, with reported errors reaching $5.8^{\circ}$ \cite{hooge2023robust,onkhar2024evaluating}. Relative motion between eye and scene cameras, inaccurate depth assumptions, and illumination-induced pupil-size changes can introduce further spatial errors of several degrees \cite{velisar2024noise,choe2016pupilSize,salari2026pupilSize}. Even systems that are comparatively robust across lighting conditions may therefore benefit from person- and session-specific offset correction \cite{baumann2026neonAccuracy,lin2025recentProgress}.

Prior correction approaches address slippage through regression models or compensate head motion in controlled environments \cite{santini2019grip,park2026motionCorrected}. Most closely, Jia et al.\ used detected traffic signs to correct gaze drift \cite{jia2018gaze}, but inferred intended targets probabilistically. In contrast, we validate wearable gaze output during real driving against explicitly declared, geometrically localized targets and relate the resulting errors to matched laboratory protocols.

\section{Methodology}
\label{sec:methodology}

\subsection{Field of View and Interior Segmentation}
\label{sec:fov}
\begin{wrapfigure}[9]{r}{0.5\textwidth}
    \vspace{-1.8\baselineskip}
    \centering
    \includegraphics[width=0.4\textwidth]{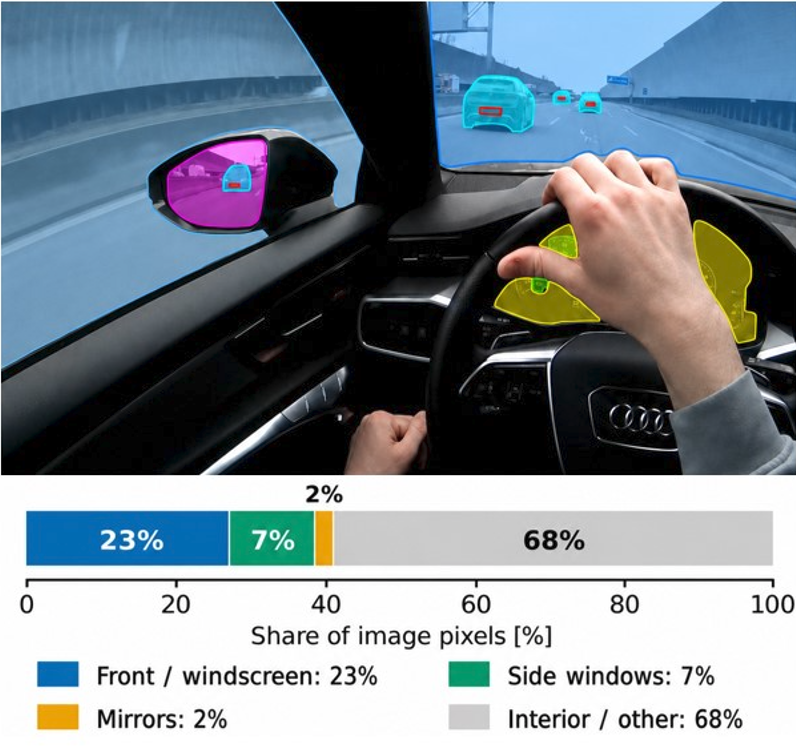}
    \caption{Scene-camera perspective.}
    \label{fig:scn_cam_perspective}
\end{wrapfigure}
Thirty minutes of natural driving were analyzed to define the gaze range for the controlled targets. Across 163 scene-camera frames, 32\% showed exterior road content and 68\% was occluded by the vehicle interior (Fig.~\ref{fig:scn_cam_perspective}). About 80\% of gaze fell within $-20^{\circ}$ to $+20^{\circ}$ azimuth and $0^{\circ}$ to $+20^{\circ}$ elevation, motivating the checkerboards' azimuth/elevation and FOV coverage.

\subsection{On-Road Gaze-Error Measurement}
\label{sec}

During driving, the driver intentionally fixated the license plate of a selected vehicle and verbally declared the target vehicle. A keyword-based speech detector located each declaration in the synchronized scene-camera audio. A YOLO-based detector provided vehicle candidates, from
which the declared target was manually selected and tracked. A second detector
localized its license plate. The plate center served as the reference because it
defines a precise fixation point, unlike the center of the full vehicle box.
All detections and fixations were manually reviewed, yielding 41 validated
scenes in which the driver fixated the license plate.

Gaze samples and target detections were synchronized, undistorted using the
camera intrinsics, and converted into unit rays. Angular gaze error was defined
as the angle between the ray toward the fixated plate center and the gaze ray
reported by the glasses. Signed azimuth and elevation errors measure systematic bias, while precision summarizes within-fixation variation in both components.

\subsection{Controlled Indoor Factor Analysis}
\label{sec:controlled-protocols}

\begin{figure}[!h]
\centering
\includegraphics[width=0.9\textwidth]{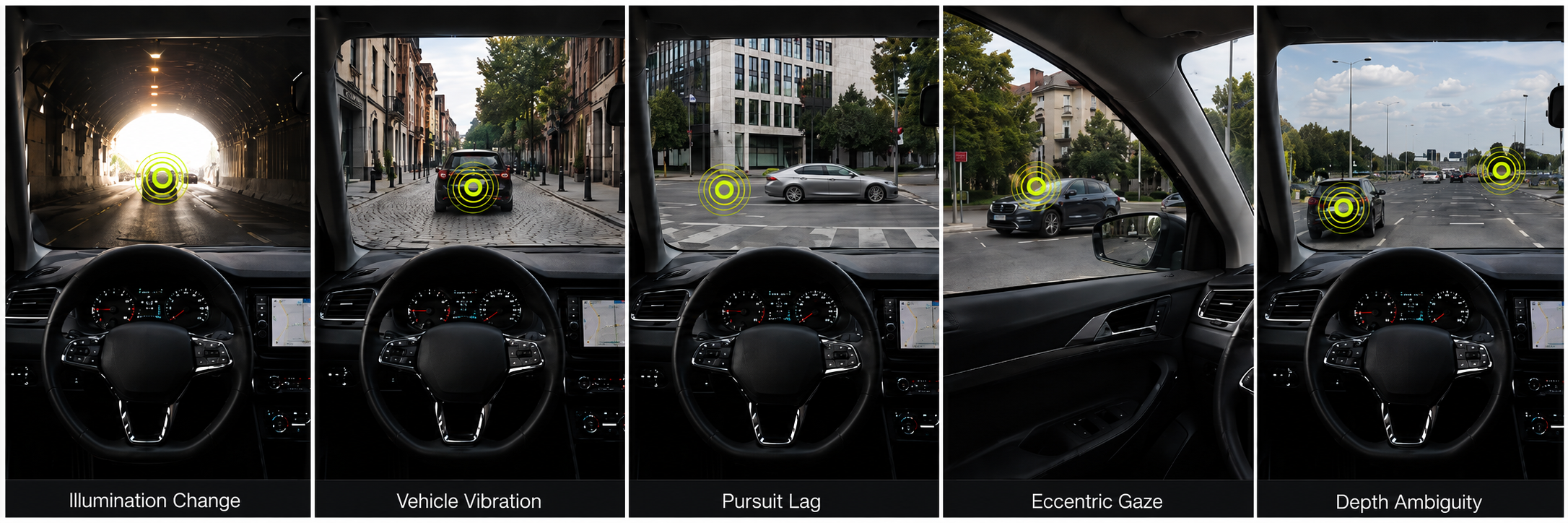}
\caption{Driving conditions motivating the protocols: illumination changes, vehicle and head motion, moving targets, eccentric gaze, and varying depth.}
\label{fig:accuracy-factors}
\end{figure}

To complement the on-road measurements in Sec.~3.2, we collected a separate indoor dataset with the same driver and eye tracker. Five controlled protocols isolate driving-relevant factors and are illustrated by their real-world manifestations in Fig.~\ref{fig:accuracy-factors}.

\noindent Each protocol uses a target designed for its measurement: \textbf{illumination} through brightness sweeps during checkerboard-corner fixation; \textbf{motion and vibration} through a circular grid during head rotations and platform vibration, whose centroids are robust to blur and rolling-shutter distortion \cite{oth2013rolling,song2025circular}; \textbf{pursuit lag} through a crosshair moving over an ArUco board at predefined speeds; \textbf{eccentric gaze} across checkerboard corners; and \textbf{depth ambiguity} through scaled checkerboards at 2--12\,m with constant angular size, with the static 2\,m condition serving as baseline. Together, they measure factor-specific changes in systematic bias and precision.

\subsection{Offset Estimation and Transfer}
\label{sec:offset-estimation}

A calibration scene's offset is the median per-fixation target-to-gaze angular error, in azimuth and elevation, in scene-camera coordinates. Transfer adds one scene's offset unchanged to another's gaze before error recomputation, excluding self-pairs. For the road result, we take the median of the six static-scene offsets and apply it to every road scene.

\section{Preliminary Results}
\label{sec:results}

This single-driver pilot includes 41 validated on-road scenes and separate indoor tests of illumination, head motion, pursuit, eccentricity, and depth. The indoor data comprise 405 target fixations and three pursuit recordings.

\begin{itemize}
    \item \textbf{System Baseline:} The static 2\,m checkerboard condition was repeated three times, with the driver fixating each of the 15 corners in every repetition. The resulting mean accuracy was 4.08\textdegree{}, while target-relative precision was 0.31\textdegree{} (median) and 0.37\textdegree{} (mean).
    
    \item \textbf{Bias Correction \& Offset:} A systematic wearing-related offset dominated the raw error. Transfer between calibration scenes improved every combination (\textbf{100\% positive transfer}; mean improvement $\sim$3\textdegree). Applying the offset to the road scenes reduced mean error from 4.58\textdegree{} to 1.10\textdegree{} (\textbf{76\%}), improving every scene.

    \item \textbf{Eccentricity:} Targets were grouped into tertiles by their angular distance from the scene-camera center. Median azimuth bias increased from approximately 0.22\textdegree{} in the central region to 0.74\textdegree{} in the outer region, indicating that a single global offset provides only a first-order correction.
    
    \item \textbf{Distance \& Speed:} Mean error ranged from 4.78--5.26\textdegree\, in the 2--12\,m sweep and pursuit medians from 4.03--4.31\textdegree; neither showed a monotonic trend.
    
    \item \textbf{Illumination:} Brightness changes increased dispersion by roughly 40--50\%.

    \item \textbf{Head Motion:} Head motion produced the largest apparent increase in camera-relative dispersion, reaching 10\textdegree; this cannot quantify tracker error because compensatory eye movements and possible slippage are not separated.
\end{itemize}

\section{Discussion and Limitations}
\label{sec:discussion}

This pilot establishes a full pipeline for measuring wearable gaze error, from controlled indoor protocols to real driving. An offset estimated purely from indoor recordings substantially reduced angular error across all 41 road scenes, and the controlled protocols show how individual driving-relevant factors affect accuracy and precision.

The next step is to extend this pipeline to more drivers, sessions, and conditions, providing enough data to model both the most likely gaze location and its uncertainty. Representing gaze as an uncertain spatial region, rather than an exact point, would support uncertainty-aware object assignment in BEV-based supervision.

\bibliographystyle{template/gcpr2026_author_kit/splncs04}
\bibliography{bib/references}

\end{document}